\documentclass[10pt,twocolumn,letterpaper]{article}

\usepackage[pagenumbers]{iccv} 

\definecolor{iccvblue}{rgb}{0.21,0.49,0.74}
\usepackage{amsmath}
\usepackage{float}
\usepackage{graphicx}
\usepackage{makecell}
\usepackage{multirow}
\usepackage[pagebackref, colorlinks=true]{hyperref}
\usepackage[linesnumbered,ruled,vlined]{algorithm2e}
\usepackage{colortbl}

\begin{document}
\title{Tracking Meets Large Multimodal Models for Driving Scenario Understanding}

\author{\textbf{Ayesha Ishaq}$^{1}$ \quad \textbf{Jean Lahoud}$^{1}$ \quad \textbf{Fahad Shahbaz Khan}$^{1,2}$ \quad \textbf{Salman Khan}$^{1,3}$ \\ \textbf{Hisham Cholakkal}$^{1}$ \quad \textbf{Rao Muhammad Anwer}$^{1}$\\
$^{1}$ Mohamed Bin Zayed University of Artificial Intelligence (MBZUAI)\\
$^{2}$ Linköping University, $^{3}$ Australian National University
}

\maketitle
\begin{abstract}

Large Multimodal Models (LMMs) have recently gained prominence in autonomous driving research, showcasing promising capabilities across various emerging benchmarks. LMMs specifically designed for this domain have demonstrated effective perception, planning, and prediction skills. However, many of these methods underutilize 3D spatial and temporal elements, relying mainly on image data. As a result, their effectiveness in dynamic driving environments is limited. We propose to integrate tracking information as an additional input to recover 3D spatial and temporal details that are not effectively captured in the images. We introduce a novel approach for embedding this tracking information into LMMs to enhance their spatiotemporal understanding of driving scenarios. By incorporating 3D tracking data through a track encoder, we enrich visual queries with crucial spatial and temporal cues while avoiding the computational overhead associated with processing lengthy video sequences or extensive 3D inputs. Moreover, we employ a self-supervised approach to pretrain the tracking encoder to provide LMMs with additional contextual information, significantly improving their performance in perception, planning, and prediction tasks for autonomous driving. 
Experimental results demonstrate the effectiveness of our approach, with a gain of 9.5\% in accuracy, an increase of 7.04 points in the ChatGPT score, and 9.4\% increase in the overall score over baseline models on DriveLM-nuScenes benchmark, along with a 3.7\% final score improvement on DriveLM-CARLA.
Our code is available at \href{https://github.com/mbzuai-oryx/TrackingMeetsLMM}{https://github.com/mbzuai-oryx/TrackingMeetsLMM}

\end{abstract}
\section{Introduction}
\label{sec:intro}

Large Multimodal Models (LMMs) have emerged as powerful tools that possess holistic knowledge to solve tasks at the intersection of vision, language, and other modalities. This has motivated their use in creating autonomous driving (AD) applications, which can enable drivers to interact with interpretable language representations of various driving safety tasks \cite{li2024driving, zhang2024chatscene}. Furthermore, LMMs can serve as end-to-end autonomous driving systems unifying traditionally separate modules, like perception and trajectory planning, into a single and cohesive framework, thereby reducing integration challenges \cite{vlp, lmdrive}. Recent advancements have led to the development of many vision-language and multimodal language models tailored for autonomous driving and the emergence of several benchmarks for Autonomous Driving Question Answering \cite{nuinstruct, drivelm, lingoQA}.

\begin{figure}
    \centering
    \includegraphics[trim=0 0 0 0,width=1\linewidth]{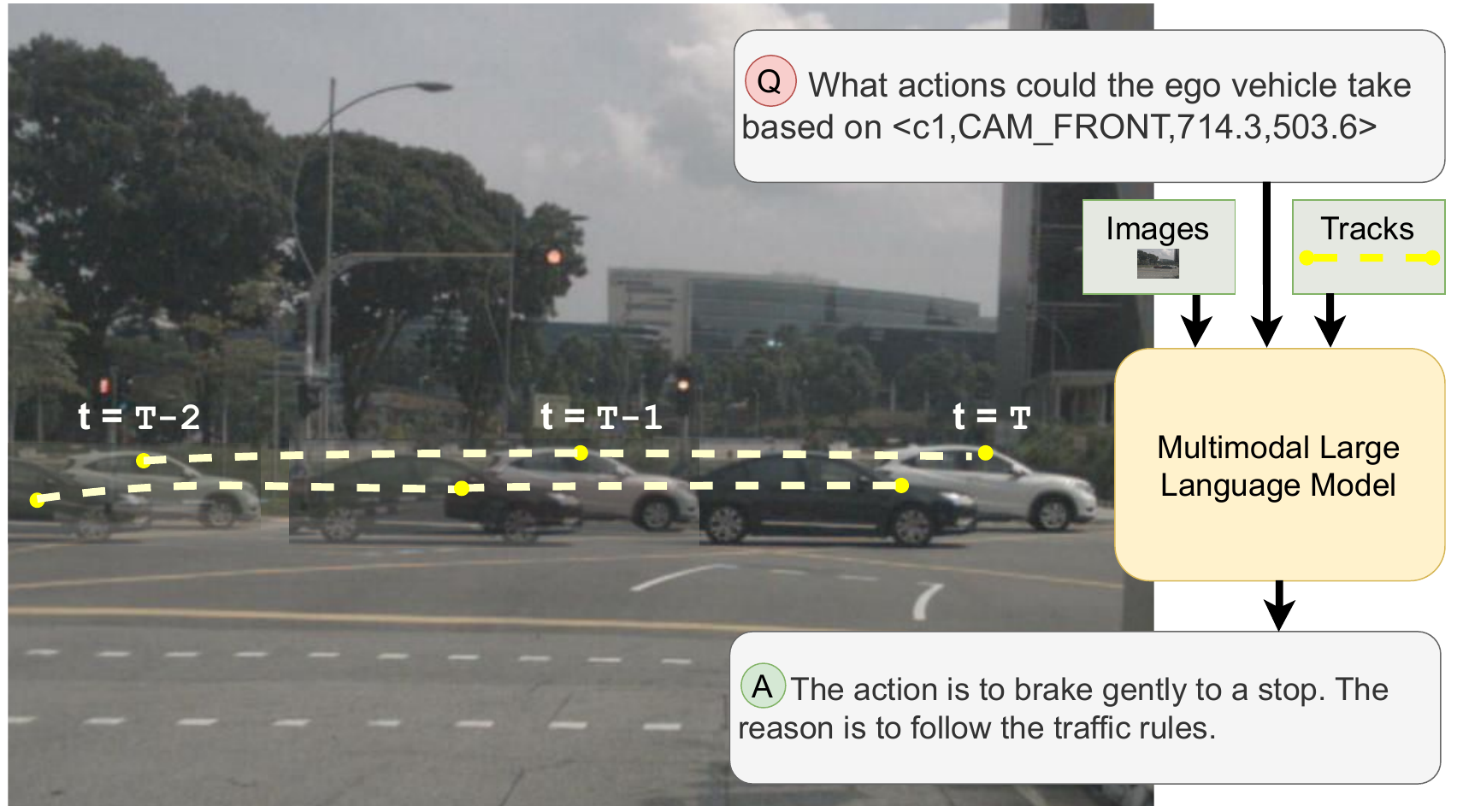}
    \caption{\textbf{Our proposed approach.} We integrate image and tracking information to enhance question-answering in the autonomous driving domain. In this example, the visual information is enriched with tracking data to provide crucial context about object movements and interactions over time. The additional tracking information allows the model to better interpret the driving scenario.}
    \label{fig:teaser}
\end{figure}

\begin{figure*}[ht]
    \centering
    \includegraphics[width=0.94\linewidth,trim=0 3mm 0 0, clip]{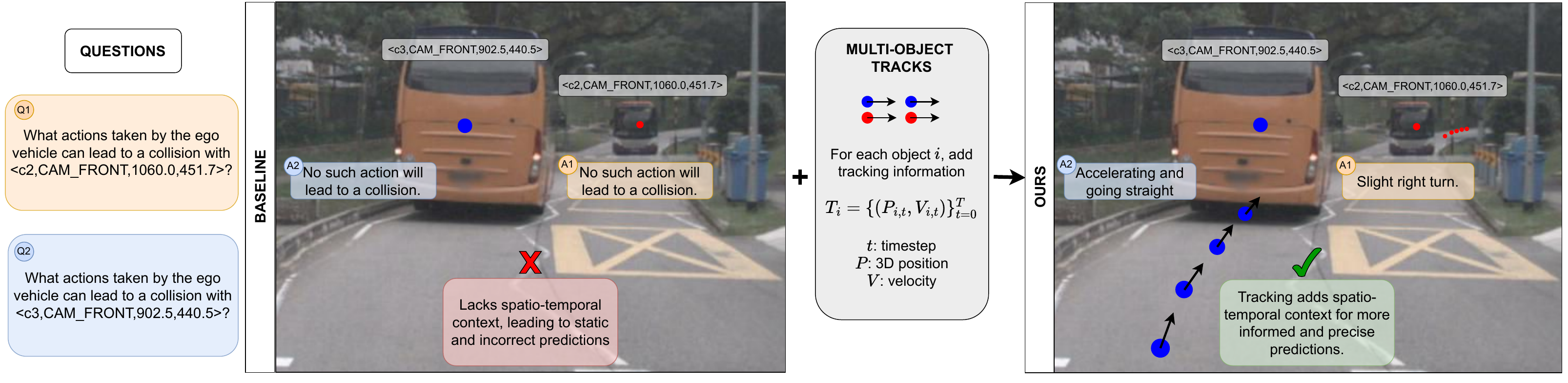}
    \caption{\textbf{
    Addressing the limitations of LMM-based methods for driving scenario understanding.} The baseline approach falls short in capturing object movements and interactions. This limitation arises from its reliance solely on image data, which lacks the necessary temporal dimension to understand dynamic environments. In contrast, our proposed method addresses these shortcomings by incorporating tracking information, including object locations and velocities, enabling the model to leverage enhanced spatiotemporal context. 
    }
    \label{fig:single_example}
\end{figure*}

Traditional LMMs have primarily focused on processing static visual inputs \cite{clip,blip, kosmos, uniter}, often overlooking essential temporal and 3D contextual information crucial for understanding dynamic environments like those found in autonomous driving scenarios. Incorporating such information, for example, through video inputs or 3D point clouds, can be computationally expensive, requiring significant resources for data processing, storage, and model training. This limitation hinders the ability of LMMs to effectively interpret evolving situations, such as the movement of multiple objects and their interactions over time, which are essential for safe and reliable decision-making in driving contexts.
Moreover, neglecting 3D spatial cues can limit the model's depth perception and understanding of complex maneuvers, as seen in Figure \ref{fig:single_example}, making it challenging to accurately predict trajectories or assess occlusions. These shortcomings highlight the need for enhanced LMMs that integrate temporal and 3D information to provide a more comprehensive understanding of driving scenarios, which is crucial for advancing autonomous driving technologies.

To address these limitations, we introduce a method to integrate tracking information into LMMs, enabling a better understanding of the spatiotemporal dynamics of driving scenarios. By incorporating tracking data, our approach enriches the model's context with vital 3D information often missing from standard image inputs. As illustrated in Figure \ref{fig:teaser}, this integration allows the LMM to interpret dynamic environments more effectively, by considering object motion and interaction over time, improving perception, planning, and prediction in autonomous driving.
In Figure \ref{fig:single_example}, an example highlights the limitations of a baseline approach \cite{drivelm}, which does not utilize crucial motion dynamics. This comparison emphasizes the advantages of our proposed method, which incorporates tracking data to enhance the understanding of dynamic environments.
Moreover, we introduce a pretraining strategy combined with an automated annotation pipeline that significantly enhances the tracking encoder's ability to understand and interpret dynamic environments. Our key contributions are as follows:

\begin{itemize}
\item We propose a tracking encoder to integrate object tracks and multimodal fusion of tracking and visual features, effectively serving as enhanced input for the Multimodal Language Model and enriching the model with spatio-temporal context.
\item To further enhance the contextual understanding of the tracking encoder, we design a self-supervised pretraining strategy and develop an automated annotation pipeline to generate pretraining data.
\item Experiments show that incorporating tracking substantially improves the model's ability to understand driving scenarios, with a gain of 9.5\% in accuracy, 7.04 points in ChatGPT score, and +9.4\% in overall score over baseline.
\end{itemize}

\begin{figure*}[t]
    \centering
    \includegraphics[trim=0 1mm 0 1mm,clip,width=1\linewidth]{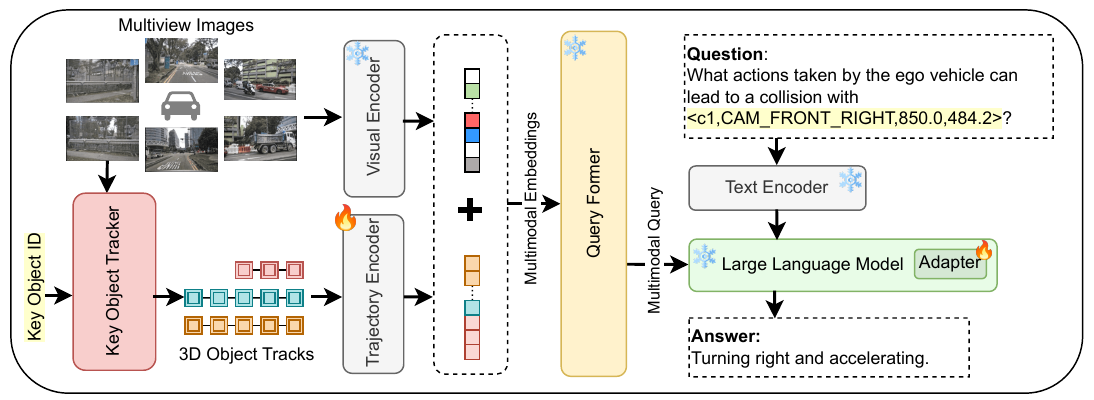}
    \caption{\textbf{Overview of our proposed method.} Our system integrates visual, trajectory, and ego-motion information through visual and trajectory encoders. 
    The input consists of multi-view images, which are processed to obtain visual embeddings, while a 3D tracker generates key object tracks. These tracks are then used to extract spatiotemporal embeddings, which are fused with the visual embeddings and transformed through a query former module.
    These multimodal embeddings are then passed to the large language model enhanced with adapters to align visual, tracking and textual modalities, enabling contextual reasoning and task-specific answers.}
    \label{fig:pipelineoverview}
\end{figure*}

\section{Related Work}
\label{sec:literature}

\paragraph{Large Multimodal Models in Autonomous Driving}
Recent advances in LMMs have demonstrated notable capabilities in autonomous driving applications. These models leverage visual and textual data to interpret complex driving environments, enabling them to perform crucial decision-making tasks. 
UP-VL \cite{najibi2023unsupervised} presents a method that utilizes vision-language models for unsupervised 3D perception tasks in autonomous driving, focusing on assigning open-vocabulary semantic labels to detected objects using VL knowledge distillation. DriveGPT4 \cite{drivegpt4} introduces an interpretable end-to-end autonomous driving system based on LLMs, which processes video inputs and textual queries to predict vehicle control signals, interpret actions, and provide reasoning. Another work, LingoQA \cite{lingoQA}, develops a VQA benchmark in autonomous driving and establishes a baseline using fine-tuned vision-language models with late video fusion for end-to-end autonomous driving tasks. 
An additional study, VLP \cite{vlp}, is a Vision Language Planning model that integrates language models into vision-based autonomous driving systems. VLP includes novel BEV reasoning and decision-making components by leveraging LMs across multiple AD system stages. 

\vspace{-3mm}
\paragraph{Methods Embedding Additional Information into LMMs}
Talk2BEV \cite{talk3bev} interfaces bird’s-eye view (BEV) maps in autonomous driving with LMM to handle various driving tasks without BEV-specific training. While the BEV adds spatial context, it lacks temporal information. Driving with LLMs \cite{drivingwithllms} presents a novel method for incorporating numeric vector modalities; compact data types such as speed, actuator positions, and distance into pretrained LLMs through adapter-based fusion. By integrating object-level 2D scene representations commonly used in autonomous driving, the approach equips LLMs to perform advanced reasoning and action prediction. 
Another method, BEV-InMLLM \cite{nuinstruct}, utilizes BEV input and enhances a novel Multi-View Multimodal LLM with BEV features for autonomous driving.
By integrating a BEV injection module with a Multi-view Q-Former, they address the limitations of existing MLLMs in capturing spatial, distance-sensitive, and occlusion-related information. A versatile model, ImageBind \cite{girdhar2023imagebind}, unifies six modalities—images, text, audio, depth, thermal, and inertial measurement unit (IMU) data—into a single embedding space, enabling cross-modal retrieval and generation tasks across all modality combinations.
However, none of these approaches incorporate tracking information.

\vspace{-3mm}
\paragraph{Multi Object Tracking in Autonomous Driving}
Recent advancements in multi-object tracking (MOT) for autonomous driving have focused on enhancing accuracy and robustness through innovative methodologies. The introduction of MCTrack \cite{ge2022mctrack} presents a unified 3D MOT framework that achieves state-of-the-art performance across multiple datasets, addressing the challenge of generalizability in existing tracking paradigms. Additionally, 3DMOTFormer \cite{ding20233dmotformer} introduces a transformer-based 3D MOT framework that uses an Edge-Augmented Graph Transformer for data association. Poly-MOT \cite{polymot} presents an innovative 3D multi-object tracking framework that dynamically selects tracking criteria tailored to different object categories, leveraging distinct motion models and a two-stage data association strategy for improved tracking. Another work, ADA-Track \cite{ding2024ada}, introduces an innovative end-to-end framework for multi-camera 3D multi-object tracking that leverages decoupled task-dependent queries with a differentiable association module. By employing an alternating optimization strategy, the method strongly couples detection and association in a more reasonable and effective manner, improving tracking accuracy. Open3DTrack \cite{ishaq2024open3dtrack} extends 3D MOT to include open-vocabulary classes, enabling the tracking of objects beyond predefined categories, thereby improving adaptability in dynamic environments.  These developments contribute to more reliable and efficient multi-object tracking systems in autonomous driving.

Building on the advancements in multimodal models, we address their limitations in capturing temporal and spatial dynamics by introducing a tracking encoder that integrates spatiotemporal object tracks. Unlike prior methods that rely on static image inputs, our approach enriches the model with 3D context through multimodal fusion of tracking and visual features. Figure \ref{fig:single_example} illustrates this improvement. A self-supervised pretraining strategy further enhances the tracking encoder's understanding, significantly improving the model's performance in perception, prediction, and decision-making for dynamic driving scenarios.

\section{Methodology}

In this section, we present our method detailing the integration of tracking information into the system and multimodal fusion with the language model. Figure \ref{fig:pipelineoverview} illustrates the pipeline of our method, along with a sample input scene, question, and expected answer.

\subsection{Baseline Architecture}
The baseline model, DriveLM \cite{drivelm}, combines a pretrained LLaMA \cite{touvron2023llama} language model with visual features from a CLIP-based \cite{clip} vision model to create a multimodal system capable of understanding both text and images. 
By leveraging their complementary strengths, 
the model is able to bridge the gap between vision and language tasks.
It enhances the language model with a lightweight adapter mechanism \cite{zhang2024llama}, allowing it to process visual inputs alongside text. This is achieved by injecting visual features as additional queries into specific layers of the language model, enabling it to reason about the data jointly.
The modular approach supports fine-tuning for domain-specific tasks.

\vspace{-3mm}
\paragraph{Baseline Limitations}
While the baseline effectively combines textual and visual modalities, it has limitations in addressing the complex demands of visual question answering (VQA) for perception, prediction, planning, and behavior understanding in autonomous driving. The reliance on pretrained CLIP features and lightweight adapters alone does not fully capture critical spatial, temporal, and depth-related information. Additionally, the lack of explicit mechanisms for modeling the interaction between objects in a 3D space limits its ability to handle multi-agent scenarios and real-time decision-making. These gaps highlight the need for a more robust integration of driving-specific features and temporal reasoning to achieve optimal performance in autonomous driving tasks.

\subsection{Tracking Information Generation}
\label{section:info_gen}

We leverage reliable state-of-the-art 3D detection and tracking frameworks to efficiently generate object detections and tracks, while performing language inference for each frame $t$. The 3D positions of each object $i$, \( P_{i,t} = (x, y, z) \), and their velocities, \( V_{i,t} = (v_x, v_y) \) are predicted for all objects in the scene to construct their trajectories, represented as \( T_i = \{(P_{i,t}, V_{i,t}) \,|\, t \in \text{time window}\} \), providing a detailed temporal and spatial view of the environment. The trajectory of the ego-vehicle, referred to as \( T_E = \{(P_{e,t}, V_{e,t}) \,|\, t \in \text{time window}\} \), is treated as a special case of \( T_i \), where \( e \) denotes the ego-vehicle.

To ensure comprehensive tracking with minimal overhead, we maintain trajectories \( T_i \) over a short time window encompassing the current frame and a few preceding frames. For each question \( q \), we identify a set of key objects at that frame \(
K = \{k_1, k_2, \ldots, k_m \}, \; m \leq 6\) with trajectories \( T_i \) near the ego-vehicle, denoted as \(K \to \{T_i \mid \text{d}(P_i(t), P_e(t)) \text{ is among the } m \text{ smallest}\} \), where $d$ is distance. If the question targets objects of interest \( S = \{s_1, s_2, \dots, s_n \}, \; n \leq 6, \; \text{where} \; s_l = (u_l, v_l) \text{ in pixel space}\), each specific object \( s_l \in S \) is matched to the relevant key object  \( k_j \in K \) where \( j \) indexes matched key object, by projecting the 3D location to image space and selecting the object with the smallest center distance. 
The trajectory information of final key objects \( K \) and the ego-vehicle’s trajectory \( T_E \) are then encoded and passed through the framework, along with the question \( q \) and multiview-images \( I \).
\begin{figure}[t]
    \centering
    \includegraphics[trim=0 3mm 0 0,width=.9\linewidth]{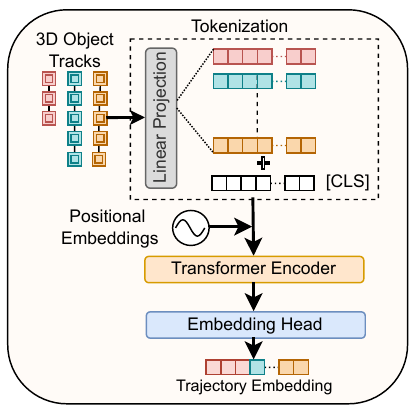}
    \caption{\textbf{Architecture of our proposed trajectory encoder.} The input tracks are tokenized via linear projection with positional embeddings. A transformer encoder refines these embeddings, capturing spatiotemporal relationships, which are output through an embedding head, preparing trajectory data for multimodal fusion.}
    \label{fig:encoder}
\end{figure}
\subsection{Integrating Tracking Information into the LMM}
\paragraph{LMM Architecture Overview}
The visual encoder within the baseline begins with a pretrained CLIP model to process input images \(I\) and extract rich visual features. The features are passed through the CLIP backbone, followed by a projection layer. The baseline language encoder utilizes a pretrained tokenizer to preprocess textual input, converting it into tokenized integer sequences, which are then passed to an embedding layer that maps each token to a high-dimensional vector representation \(L_E\).

\paragraph{Trajectory Encoder} 

Motivated by previous multimodal language models \cite{girdhar2023imagebind, talk3bev, zhang2023metatransformer}, we input object tracks as an additional modality to our multimodal system and extract embeddings from this input using an additional encoder module.
The trajectory encoder preprocesses spatiotemporal data into embeddings, denoted as \( \mathbf{E}_i \). Given a trajectory input \( T_i = \{(P_{i,t}, V_{i,t}) \,|\, t \in \text{time window}\} \), the positions and velocities over all timestamps \( t \in \text{time window} \) are flattened into a single vector:
\[
\mathbf{v}_i = [P_{i,t_1}, V_{i,t_1}, P_{i,t_2}, V_{i,t_2}, \dots, P_{i,t_n}, V_{i,t_n}],
\]
We then apply a linear projection \( f_{\text{proj}} \) to map the flattened input vector into a higher-dimensional embedding space:
\[
\mathbf{z}_i = f_{\text{proj}}(\mathbf{v}_i) + \mathbf{p}_i,
\]
where \( \mathbf{p}_i \) is the learnable positional embedding for the \( i \)-th object, capturing its temporal-spatial position within the scene.
Layer normalization \( f_{\text{norm}} \) is subsequently applied to the projected embedding 
\(
\mathbf{z}'_i = f_{\text{norm}}(\mathbf{z}_i).
\)
A class token \( \mathbf{c}_i \) is then appended to this embedding to encode global contextual information 
\(
\mathbf{Z}_i = [\mathbf{c}_i; \mathbf{z}'_i],
\)
A simple transformer encoder, \( f_{\text{enc}} \), refines this input sequence to extract semantic information and help transform this modality to language embedding space, using multi-head self-attention (MSA) and feed-forward networks (FFN):
\[
\mathbf{E}_i = f_{\text{enc}}(\mathbf{Z}_i) = \text{MSA}(\mathbf{Z}_i) + f_{\text{FFN}}(\text{MSA}(\mathbf{Z}_i)).
\]
For the ego-vehicle, a similar process is applied to its trajectory \( E \):
\[
\mathbf{E}_\text{ego} = f_{\text{enc}}(\mathbf{Z}_\text{ego}),
\]
The refined embeddings of the ego-vehicle and key objects are processed through the head module \( f_{\text{head}} \) to produce output embeddings ready for multimodal fusion \( \mathbf{O}_E \) and \( \mathbf{O} \) respectively:
\[
\mathbf{O}_E = f_{\text{head}}(\mathbf{E}_\text{ego}) , \quad \mathbf{O} = f_{\text{head}}([\mathbf{E}_1; \mathbf{E}_2; \dots; \mathbf{E}_k]),
\]

Figure \ref{fig:encoder} visualizes the structure of the trajectory encoder, showing how tracks are processed through each module.

\paragraph{Multi-modality Fusion}
The query former module (Figure \ref{fig:pipelineoverview}) integrates trajectory, ego-vehicle motion, and visual data into a unified representation. The outputs of the trajectory encoder, \( \mathbf{O} \) and \( \mathbf{O}_E \), are first normalized and projected into a shared embedding space using dedicated linear layers \( f_{\text{traj-proj}} \) and \( f_{\text{ego-proj}} \). This step ensures alignment with CLIP image features:
\[
\mathbf{F}_{\text{traj}} = f_{\text{traj-proj}}(\mathbf{O}),
\quad \mathbf{F}_{\text{ego}} = f_{\text{ego-proj}}(\mathbf{O}_E).
\]
Each modality's feature embedding is scaled using modality-specific weights \( w_{\text{traj}} \), \( w_{\text{ego}} \), and \( w_{\text{clip}} \), provided in the input, which allow prioritizing relevant modalities. The embeddings are then summed to create a fused embedding capturing visual semantics, temporal and spatial context:  
\[
\mathbf{F}_{\text{fused}} = w_{\text{traj}} \cdot \mathbf{F}_{\text{traj}} + w_{\text{ego}} \cdot \mathbf{F}_{\text{ego}} + w_{\text{clip}} \cdot \mathbf{F}_{\text{clip}}
\]
This fused representation \( \mathbf{F}_{\text{fused}} \) is concatenated with learnable visual query embeddings \( \mathbf{Q}_{\text{visual}} \) to form the input for the transformer-based visual blocks, where each block \( f_{\text{visual-block}} \) learns the query vectors through multi-head self-attention (MSA) and feed-forward networks (FFN), capturing dependencies and interactions between the modalities:
\[
\mathbf{F}_{\text{output}} = f_{\text{visual-block}}([\mathbf{Q}_{\text{visual}}; \mathbf{F}_{\text{fused}}]).
\]
A final linear projection on the extracted queries ensures compatibility with the pretrained language model, facilitating efficient integration:
\[
\mathbf{F}_{\text{final}} = f_{\text{visual-proj-norm}}(f_{\text{visual-proj}}(\mathbf{F}_{\text{output}}[ :\text{len}(\mathbf{Q}_{\text{visual}})]). 
\]

The output query \(\mathbf{F}_{\text{final}}\) is combined with the adapter query, a set of learnable embeddings designed to inject task-specific multimodal context into the language model layers. The adapter query enables fine-tuning specific layers for multimodal alignment without modifying or retraining the entire language model. This approach preserves the efficiency and generalization capabilities of the pretrained model while adapting it effectively to the task of visual question answering with fused modalities.
\begin{figure}
    \centering
    \includegraphics[trim=0 2mm 0 1mm,clip,width=1\linewidth]{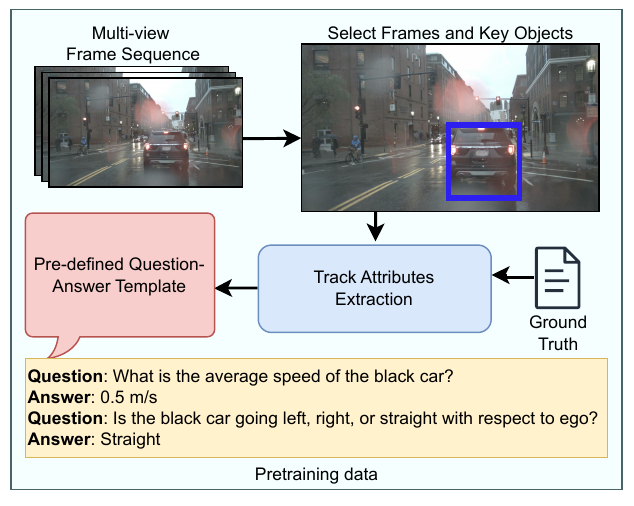}
    \caption{\textbf{Overview of the automated annotation pipeline.} We make use of nuScenes \cite{nuscenes2019} ground truth and multi-view frames to generate tracks and question-answer pairs related to track attributes for pretraining our proposed trajectory encoders.}
    \label{fig:pretraining}
\end{figure}

\paragraph{Pretraining the Tracking Encoder}
To allow the model to learn representative patterns and capture useful context from the trajectory inputs, we pretrain the trajectory and ego-vehicle trajectory encoder instances on large-scale data generated by a fully automated pipeline by leveraging the nuScenes \cite{nuscenes2019} dataset. The process involves selecting key frames and objects of interest within a scene and extracting their attributes to create ground truth information. We define question-answer templates, focusing on extracting meaningful trajectory-related insights. For example, questions like "What is the average speed of the black car?" and "Is the pedestrian going left, right, or straight with respect to ego-vehicle?" are answered based on trajectory and object attributes derived from the dataset ground truth. The method processes the trajectories of the ego-vehicle and other objects by tracking their positions over time, calculating average speed and acceleration status, and predicting future positions. Additionally, bounding boxes are extracted to identify objects in images, and their dominant colours are computed for generating more descriptive object references.
Figure \ref{fig:pretraining} illustrates this workflow.

\paragraph{Implementation Details}
To implement our approach, we use the 3D object detector CenterPoint \cite{centerpoint} and the tracking-by-detection framework 3DMOTFormer \cite{ding20233dmotformer} to generate reliable 3D tracks. For tracks of key objects input to the multimodal model, we limit the length to 5 frames and filter out tracks with a confidence score below 0.3. Further preprocessing involves class-agnostic Non-Maximum Suppression (NMS) in 2D with an IoU threshold of 0.1, retaining up to 6 tracks closest to the ego-vehicle for each question. 
Similar to the baseline, DriveLM \cite{drivelm}, we utilize the multimodal LLaMA-Adapter-v2 \cite{zhang2024llama} based on LLaMA-2 7B \cite{touvron2023llama}. The visual encoder integrates a pretrained CLIP \cite{clip} model (ViT-L/14) with projection layers trained on LAION-400M \cite{schuhmann2021laion} and zero-initialized attention fine-tuned on data from Alpaca and LLaVA-I \cite{liu2024visual}. $w_{\text{traj}}, w_{\text{ego}}, w_{\text{clip}}$ in modality fusion are all set to 1. Fine-tuning is conducted with a warmup epoch followed by three additional epochs, using a base learning rate of $10e^{-4}$, weight decay of 0.02, and half-cycle cosine learning rate scheduling after the warmup phase. For pretraining, 5 warmup epochs followed by 145 epochs, a base learning rate of $1e^{-4}$, weight decay of 0.05, and the same learning rate scheduler are used. We conduct all experiments on 2 RTX A6000 GPUs. 
\section{Experiments}

\begin{table*}[h]
\centering
\renewcommand{\arraystretch}{1.5} 
\resizebox{0.9\textwidth}{!}{%
\begin{tabular}{c|lccccccccc|c}
\hline
& \textbf{Method}               & \textbf{Accuracy} & \textbf{ChatGPT} & \textbf{Bleu 1}  & \textbf{Bleu 2} & \textbf{Bleu 3} & \textbf{Bleu 4} & \textbf{ROUGE L} & \textbf{CIDEr} & \textbf{Match}  & \textbf{Final Score} \\ \hline

 & DriveLM \cite{drivelm}            & 0.454    & 55.272  & 0.680  & 0.625  &  0.566  & 0.512  & 0.672   & 0.006 & 33.750 & 0.464     \\
& DriveLM + Video & 0.667    & 56.000  & 0.670  & 0.606  & 0.542  & 0.485  & 0.676   & 0.014 & 39.250 & 0.519       \\
\rowcolor{gray!10}
\multirow{-3}{*}{\rotatebox{90}
{\cellcolor{white} Val Set}}
& \textbf{Ours} & \textbf{0.818}    & \textbf{67.090}  & \textbf{0.730}  & \textbf{0.670}  & \textbf{0.610}  & \textbf{0.552}  & \textbf{0.734}   & \textbf{0.140} & \textbf{43.500}&  \textbf{0.611}  \\ \hline

 & DriveLM \cite{drivelm}     & 0.501    & 51.396  & 0.614  & 0.559  & 0.506  & 0.455  & 0.658   & 0.033 & 17.883 & 0.421     \\
& DriveLM + Video & 0.523    & 55.352  & 0.706  & 0.642  & 0.581  & 0.522  & 0.701  & 0.040 & 29.074 & 0.472      \\
\rowcolor{gray!10}
\multirow{-3}{*}{\rotatebox{90}
{\cellcolor{white} Test Set}}
& \textbf{Ours}                 & \textbf{0.596}    & \textbf{58.438}  & \textbf{0.724}  & \textbf{0.658}  & \textbf{0.593}  & \textbf{0.531}  & \textbf{0.721}   & \textbf{0.044} & \textbf{35.730}&  \textbf{0.515}       \\ \hline

\end{tabular}%
}
\caption{\textbf{Results on DriveLM Benchmark.} Results are reported on DriveLM-nuScenes test and validation set. We use the sample set from DriveLM as validation set. Baseline results are obtained from fine-tuning DriveLM \cite{drivelm} baseline on DriveLM-nuScenes train set. Our method achieves better performance compared to baseline and baseline with video input.}
\label{tab:mainresults}
\end{table*}

\begin{table}[t]
\centering
\renewcommand{\arraystretch}{1.1} 
\resizebox{0.48\textwidth}{!}{%
\begin{tabular}{p{1.5cm}llllll}
\hline
\textbf{Method} & \textbf{Accuracy} & \textbf{ChatGPT} & \textbf{Bleu 1} & \textbf{ROUGE L} & \textbf{CIDEr} & \makecell{\textbf{Final} \\ \textbf{Score}} \\ \hline
DriveLM \cite{drivelm} & 0.908 & 72.824   & 0.732 & 0.712 & 1.145 &  0.714 \\
\textbf{Ours} & \textbf{0.917} & \textbf{77.972} & \textbf{0.753} &\textbf{0.736 }& \textbf{1.617} &  \textbf{0.751} \\ \hline
\end{tabular}%
}
\caption{\textbf{Results on DriveLM-CARLA.} Results are reported for a subset of CARLA Town13 used as test set. 
The same baseline from DriveLM-nuScenes has been used here. 
}
\label{tab:carla}
\end{table}

\subsection{Dataset}
We use the DriveLM dataset \cite{drivelm}, designed to support autonomous driving tasks with detailed annotations on perception, prediction, and planning. The annotation process includes selecting keyframes, identifying key objects, and generating QA pairs for these objects. For DriveLM-nuScenes, perception QAs are partially derived from nuScenes \cite{nuscenes2019} and OpenLane-V2 \cite{wang2023openlanev2} ground truth, while others are manually annotated using templates crafted by domain experts. However, DriveLM-CARLA VQAs are automatically generated using a rule-based approach. DriveLM-nuScenes includes 4,871 frames with an average of 91.4 QAs per frame, while DriveLM-CARLA features 183k frames with an average of 20.4 QAs per frame. This large-scale and high-quality dataset offers comprehensive coverage of autonomous driving scenarios, including real-world and simulation scenes, making it a valuable resource for advancing autonomous systems.

\vspace{-3mm}
\paragraph{Evaluation Metrics}
We employ the DriveLM \cite{drivelm} benchmark to evaluate our results. This evaluation framework uses multiple metrics to assess system performance across various aspects. Accuracy is calculated as the proportion of predictions for multiple choice and yes/no questions that match the ground truth answers. ChatGPT score evaluates the semantic alignment of the responses by prompting GPT-3.5 Turbo \cite{openai2023gpt3.5turbo} to access the similarity in the generated and ground truth answer. Language metrics, including BLEU \cite{papineni2002bleu}, ROUGE-L \cite{lin2004rouge}, and CIDEr \cite{vedantam2015cider}, evaluate linguistic quality and similarity between generated and ground truth answers. A matching score is computed by evaluating the F1-score of object 2D location matching between predictions and ground truth. The final composite score is derived by normalizing these metrics to a 0-1 scale and combining them using predefined weights, with ChatGPT score contributing the most.

\subsection{Results}
\paragraph{DriveLM-nuScenes}
We show the results of our method in comparison to the baseline in Table \ref{tab:mainresults}. For training, we use the DriveLM-nuScenes train set, excluding the sample data that we use as our validation set. We evaluate on both DriveLM-nuScenes test set and the validation set.
The results highlight the effectiveness of our proposed method, which achieves the highest accuracy and final score on both test (0.596 accuracy and 0.515 final score) and validation sets (0.818 accuracy and 0.612 final score), outperforming both the baseline and the baseline with video input. Unlike the baseline, which relies solely on input from static images and lacks temporal modeling, and the baseline with video input, which processes 5-frame multi-view image videos but fails to fully capture temporal relationships, our method excels in capturing nuanced temporal and spatial context with the tracking input. This is evident in the validation set, where our method achieves the highest CIDEr score (0.140) and strong match score (43.5), as well as the test set, where it improves the match score (35.730) and CIDEr score (0.044), indicating superior alignment with ground truth. Additionally, our method demonstrates strong language generation performance, as shown by the highest ChatGPT score (67.09 on validation and 58.438 on test) and competitive ROUGE-L (0.734 on validation, 0.721 on test) and Bleu-1 (0.730 on validation, 0.724 on test) scores.

\begin{figure*}[ht]
    \centering
    \includegraphics[trim=0 0 0 0,width=1\linewidth]{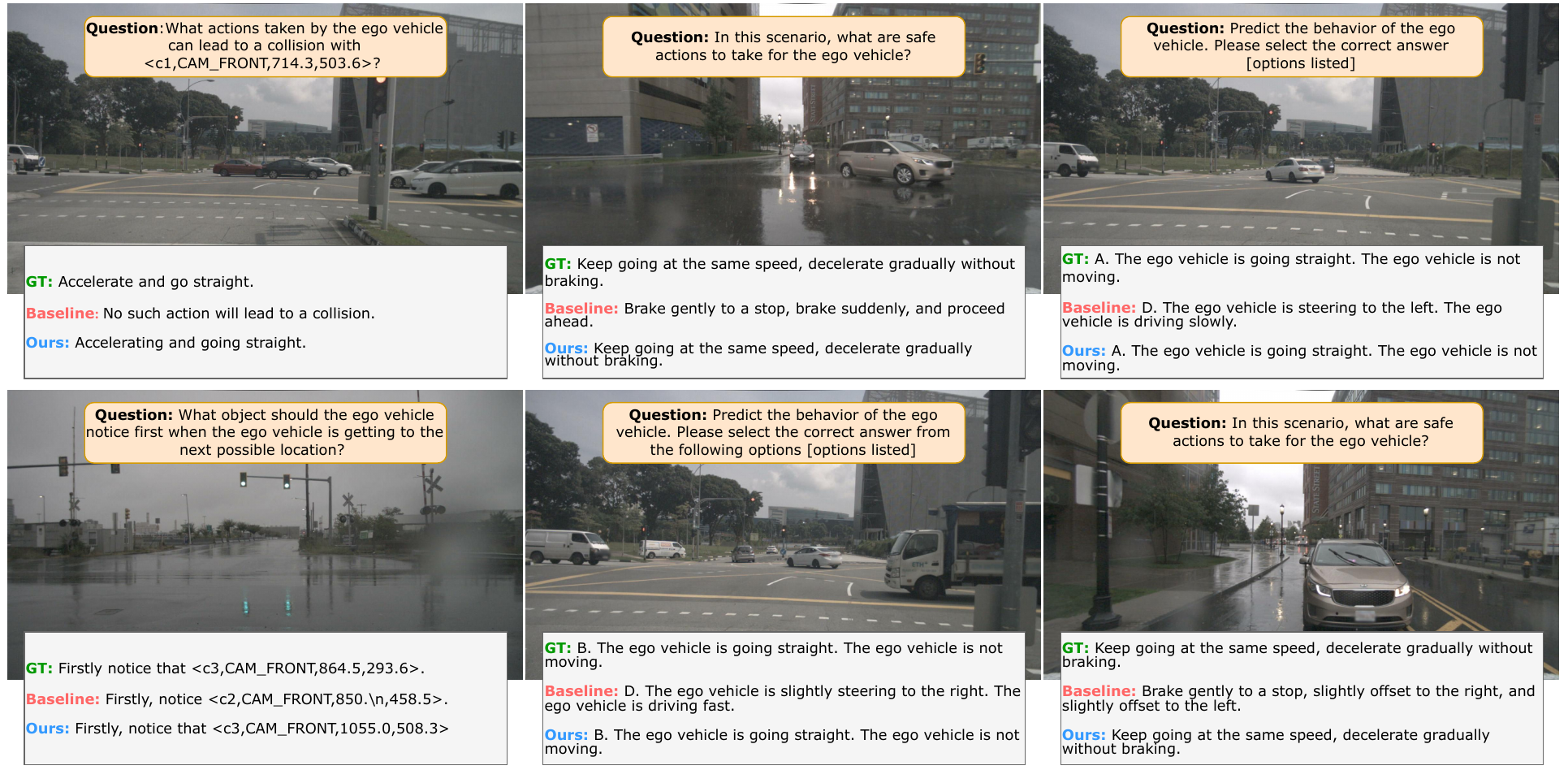}
    \caption{\textbf{Qualitative results} on the DriveLM-nuScenes sampled validation set, demonstrating our model's ability to answer mixed questions across perception, prediction, and planning tasks with superior accuracy compared to baseline. While our model excels in most tasks, a few complex cases requiring nuanced predictions or fine-grained awareness may still pose challenges. These results emphasize the robustness and versatility of our approach while also pointing to areas for potential improvement in subtle edge cases.
    }
    \label{fig:qualitative}
\end{figure*}

Our approach's ability to better model spatiotemporal dependencies across frames sets it apart, enabling accurate predictions of ego-vehicle behavior and object interactions. While the baseline with video input improves over static input, it remains limited by shallow temporal modeling. In contrast, our method processes motion and scene semantics more comprehensively, significantly improving accuracy and contextual understanding. These results validate the robustness and adaptability of our approach in handling complex multimodal inputs. Figure \ref{fig:qualitative} shows qualitative results from our validation holdout set for questions for perception, prediction, planning, and behavior tasks.

\vspace{-3mm}
\paragraph{DriveLM-CARLA}
We additionally evaluate our method on the DriveLM-CARLA.  
We report the results in Table \ref{tab:carla}.
We train the model on the Town12 subset of DriveLM-CARLA, and use Town13 as the test set. These experiments help assess our method's generalization to different driving environments. By incorporating trajectory input, we achieve 91.7\% accuracy and a 0.751 final score, outperforming the DriveLM baseline, which has a final score of 0.714. 
These results further underscore the method's robustness in various driving scenarios and highlight its potential for a wider application. Qualitative results are reported in the supplementary material. 

\vspace{-3mm}
\paragraph{Runtime} In addition to achieving superior performance, our method demonstrates competitive runtime efficiency. For each LMM prompt, the baseline processes in approximately 3.72 sec, while the baseline with multi-frame input takes 7.44 sec due to the added complexity of handling multi-frame sequences. In comparison, our method processes in approximately 3.80 sec, which includes 3.72 sec for the computation by the LMM; additional trajectory processing has no significant impact on runtime, 0.0625 sec for detection, and 0.018 sec for tracking for 5 frames (Centrepoint \cite{centerpoint} 16 FPS, 3DMOTFormer \cite{ding20233dmotformer} 54.7 Hz). Despite the additional steps for detection and tracking, our runtime remains close to the baseline, showcasing the efficiency of our multimodal integration. 

\subsection{Ablation Study}
\begin{table}[t]
\centering
\renewcommand{\arraystretch}{1.5} 
\setlength{\tabcolsep}{2pt}
\resizebox{0.48\textwidth}{!}{%
\begin{tabular}{lcccccccc}
\hline
\textbf{Encoders} & \textbf{Tracks} & \textbf{Accuracy} & \textbf{ChatGPT} & \textbf{Bleu 1} & \textbf{ROUGE L} & \textbf{CIDEr} & \textbf{Match} & \makecell{\textbf{Final} \\ \textbf{Score}} \\ \hline
Single & Object  & 0.364 & 56.455 & 0.719 & 0.742 & 0.210 & 36.625 & 0.465 \\ 
Single & Object \& Ego & 0.364 & 60.364 & 0.719 & 0.742 & 0.210 & 31.500 & 0.470 \\ 
Separate & Object \& Ego & 0.667 & 57.000 & 0.746 & 0.741 & 0.308 & 41.875 & 0.541 \\ \hline
\end{tabular}
}
\caption{Comparison of different trajectory encoder and input data configurations on val set. \textbf{Single} denotes one trajectory encoder, while \textbf{separate} denotes two encoder instances.}
\label{tab:encoders}
\end{table}

We analyze the impact of different design choices by conducting various ablation studies. We perform all ablations on a validation set which we subset from the DriveLM-nuScenes train set. We use the remaining data for training.

In Table \ref{tab:encoders}, the results of the impact of different encoder configurations on performance are presented. When a single encoder was used with tracks of only key objects during both training and testing, the model achieves an accuracy score of 0.364 and a final score of 0.465. The performance improves slightly when a single encoder was used to process both key object and ego-vehicle tracks during training and testing. The accuracy score remains at 0.364, but the ChatGPT score increases to 60.364, suggesting better alignment of generated text with ground truth. The final score also increases marginally to 0.470, indicating that incorporating ego-vehicle trajectories added complementary information. The highest performance was observed when separate encoders were used to process ego-vehicle and object tracks during training and testing. This configuration achieves the best accuracy of 0.667 and a final score of 0.541. 
These results suggest that separate encoders allow the model to better capture the unique contextual contributions of ego-vehicle and object tracks.

\begin{table}[t]
\centering
\renewcommand{\arraystretch}{1.1} 
\setlength{\tabcolsep}{3pt}
\resizebox{0.48\textwidth}{!}{%
\begin{tabular}{lccccccc}
\hline
\textbf{Methodology} & \textbf{Accuracy} & \textbf{ChatGPT} & \textbf{Bleu 1} & \textbf{ROUGE L} & \textbf{CIDEr} & \textbf{Match} & \makecell{\textbf{Final} \\ \textbf{Score}} \\ \hline
No pretraining & 0.667 & 57.000   & 0.746 & 0.741 & 0.308 & 41.875 & 0.541 \\
\makecell[l]{Pretrained trajectory \\ encoder} & 0.778 & 64.444 & 0.733 & 0.718 & 0.318 & 35.250 & 0.577 \\
\makecell[l]{Pretrained trajectory \\ and ego encoder} & 0.889 & 61.333 & 0.707 & 0.721 & 0.107 & 41.500 & 0.596 \\ \hline
\end{tabular}%
}
\caption{Ablation study results comparing effects of pretraining trajectory and ego-vehicle trajectory encoders on val set.}
\label{tab:ablation_study_pretrain}
\end{table}

The second ablation study, shown in Table \ref{tab:ablation_study_pretrain}, highlights the role of pretraining in improving system performance. Without pretraining, the system achieves an accuracy of 0.667 and a final score of 0.541. However, incorporating a pretrained trajectory encoder boosts accuracy to 0.778 and improves the final score to 0.577, indicating that pretraining on trajectory data helps the model better capture spatial-temporal patterns. When both the trajectory and ego-vehicle trajectory encoders are pretrained, the performance improves further, with accuracy rising to 0.889 and the final score improving to 0.596. Emphasizing the importance of including ego-centric dynamics in the pretraining. Metrics like ChatGPT score and BLEU 1 show a slight drop when both encoders are pretrained. 
This could be due to the limited variety of question-answer templates in the pretraining data, which results in increased specificity in the model.

In Table \ref{tab:ablation_study_prob}, we investigate the impact of mixing trajectory training data on the model performance. 
During training, ground truth (GT) tracks are randomly mixed with tracks from a 3D tracker at varying probabilities to evaluate the effects of introducing noise and data diversity in the training data. The results demonstrate that this approach enhances the model's performance by introducing valuable variety, with a 25\% mixing probability achieving the best balance between accuracy and reliability. This combination increases the final score from 0.541 (GT only) to 0.586, showcasing the advantage of incorporating tracker data. Pretraining further amplifies these benefits, achieving the highest accuracy (0.818) and final score (0.612), indicating improved robustness to noisy inputs. While higher mixing probabilities (50\%) introduce excessive noise, lowering the CIDEr score, a 25\% mix maintains precision while leveraging the benefits of tracker data.

\begin{table}[t]
\centering
\renewcommand{\arraystretch}{1.5} 
\resizebox{0.48\textwidth}{!}{%
\begin{tabular}{p{1cm}llccccccc}
\hline
\textbf{Data} & \textbf{Prob} & \textbf{Accuracy} & \textbf{ChatGPT} & \textbf{Bleu 1} & \textbf{ROUGE L} & \textbf{CIDEr} & \textbf{Match} & \makecell{\textbf{Final} \\ \textbf{Score}} \\ \hline
GT  & -    & 0.667 & 57.000   & 0.746 & 0.741 & 0.308 & 41.875 & 0.541 \\
Mix               & 0.50  & 0.778 & 61.222 & 0.757 & 0.732 & 0.034 & 41.250  & 0.576 \\
Mix               & 0.25 & 0.778 & 60.444 & 0.777 & 0.762 & 0.376 & 44.792 & 0.586 \\ 
Mix$^*$              & 0.25 & 0.818 & 67.090 & 0.730 & 0.734 & 0.140 & 43.500 & 0.612 \\ \hline
\end{tabular}%
}
\caption{
Ablation study on different dataset configurations for the val set.
\textbf{Prob} column denotes the mixing probability of additional 3DMOTFormer tracks with GT. $^*$denotes pretrained encoders.}
\label{tab:ablation_study_prob}
\end{table}

\section{Conclusion}
We introduce a novel method to integrate tracking into LMMs, enhancing their capability to interpret dynamic driving scenarios by incorporating spatiotemporal cues from object and ego-vehicle trajectories. Our approach enhances the LMM's contextual understanding by leveraging a trajectory encoder and multimodal fusion, enabling significant improvements in perception, planning, and prediction tasks. Furthermore, a pretraining strategy for trajectory encoders allows the model to capture critical spatial and temporal patterns, further boosting performance.
Experiments on the DriveLM benchmark show the merits of our method, outperforming baselines in accuracy and language generation while maintaining competitive runtime efficiency. This work provides a robust framework for multimodal integration, paving the way for advances in autonomous driving.

{
    \small
    \bibliographystyle{ieeenat_fullname}
    \bibliography{main}
}

\clearpage
\section*{Supplementary Material}









\begin{algorithm}[h]
\caption{Automated QA Generation for Pretraining}
\label{alg:data}
\KwData{NuScenes Dataset}
\KwResult{JSON file with object and ego QA data}
\Begin{
    Load NuScenes Dataset\;
    Get scene list from dataset\;
    Initialize \texttt{output} as empty\;

    \ForEach{scene in scenes}{
        \ForEach{key sample in scene}{
            Initialize frame data\;
            Get ego pose and trajectory\;
            Calculate ego attributes:
                \begin{itemize}
                    \item Average speed
                    \item Acceleration status
                    \item Predicted future position
                \end{itemize}
            Select key objects near ego vehicle\;
            Get camera images\;

            \ForEach{key object in frame}{
                Extract object trajectory for time window\;
                Calculate:
                \begin{itemize}
                    \item Average speed
                    \item Acceleration status
                    \item Predicted future position
                \end{itemize}
                Find relative direction to ego\;
                Extract object color and location from images\;
                Generate object description\;

                Append QAs for the object along with images and trajectories to \texttt{output}\;
            }

            Append QAs for ego along with images and trajectory to \texttt{output}\;
        }
    
    }

    Save \texttt{output} to JSON file\;
}
\end{algorithm}

\section*{Template QAs for Object Trajectories}

\begin{enumerate}
    \item \textbf{Question:} What is the average speed of the \textit{\textless object \textgreater} along this trajectory? \\
          \textbf{Answer:} \textit{\textless Calculated average speed in m/s \textgreater}.

    \item \textbf{Question:} Is the \textit{\textless object \textgreater} accelerating, decelerating, or maintaining a steady speed along its path? \\
          \textbf{Answer:} \textit{\textless Acceleration status \textgreater}.

    \item \textbf{Question:} Is the \textit{\textless object \textgreater} going left, right, or straight with respect to the ego vehicle? \\
          \textbf{Answer:} \textit{\textless Relative direction \textgreater}.

    \item \textbf{Question:} What is the predicted position of the \textit{\textless object \textgreater} after the last observed point? \\
          \textbf{Answer:} \textit{\textless Predicted x, y position \textgreater}.
\end{enumerate}

\section*{Template QA for Ego-Vehicle Trajectory}
\begin{enumerate}
    \item \textbf{Question:} What is the average speed of the ego vehicle along this trajectory? \\
          \textbf{Answer:} \textit{\textless Average speed of the ego vehicle \textgreater}.

    \item \textbf{Question:} Is the ego vehicle accelerating, decelerating, or maintaining a steady speed along its path? \\
          \textbf{Answer:} \textit{\textless Acceleration status \textgreater}.

    \item \textbf{Question:} What is the predicted position of the ego vehicle after the last observed point? \\
          \textbf{Answer:} \textit{\textless Predicted x, y position \textgreater}.
\end{enumerate}
\vspace{0.6cm} 
The pretraining data generation process, as detailed in Algorithm \ref{alg:data}, extracts spatiotemporal trajectories of the ego vehicle and surrounding objects alongside visual data from multiple cameras. The template captures key attributes like average speed, acceleration status, future positions, relative directions, and object descriptions (color and location), represented as targeted QA pairs. This design ensures that dynamic interactions and spatial relationships are well-represented. Pretraining on this data equips the model with enhanced contextual understanding and predictive capabilities, improving downstream tasks in autonomous driving, including perception, planning, and decision-making.



\begin{figure*}
    \centering
    \includegraphics[width=0.9\linewidth]{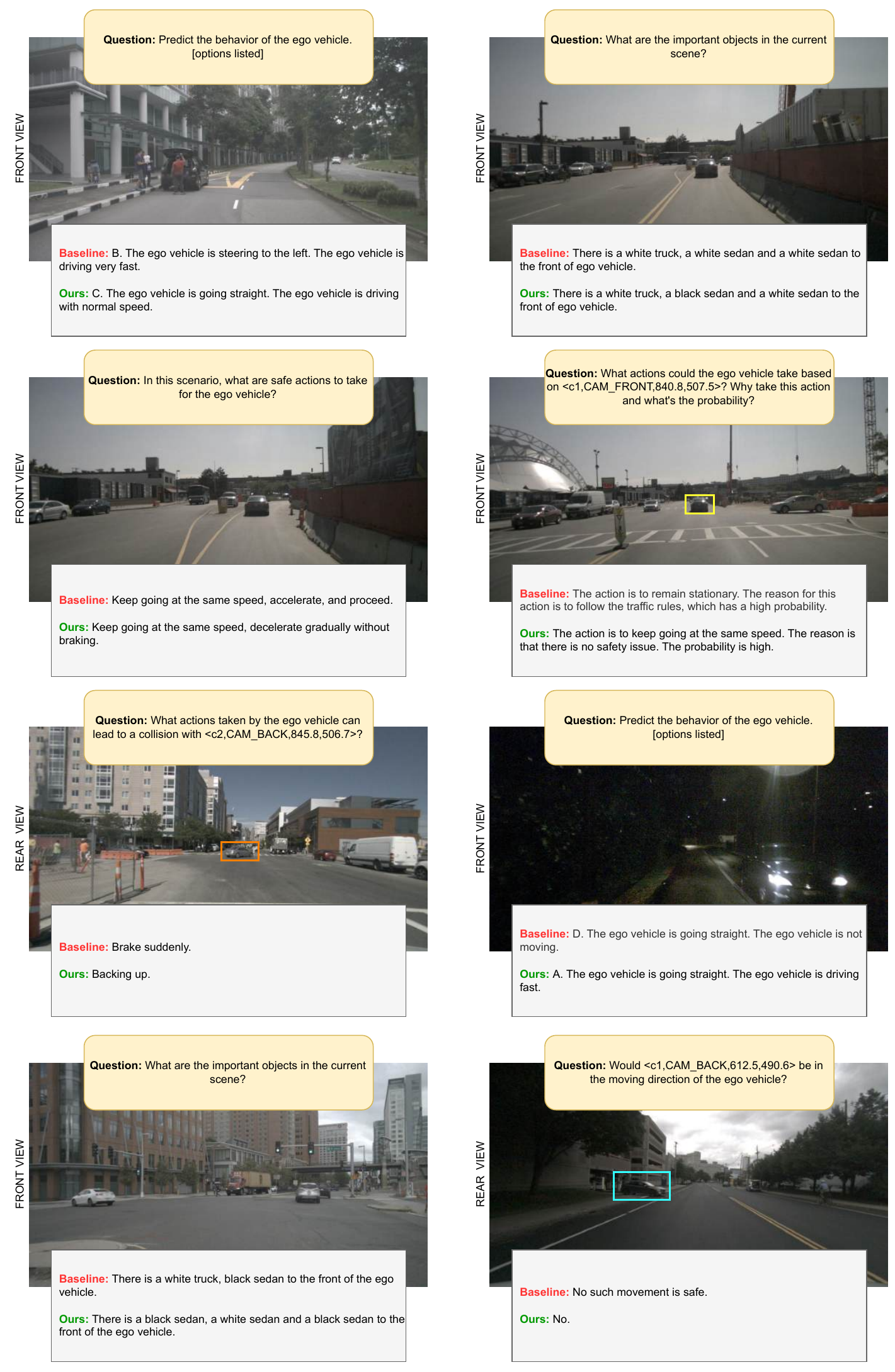}
    \caption{\textbf{Qualitative Results from DriveLM-nuScenes test set.} These examples present a comparison between baseline and our results.}
    \label{fig:enter-label}
\end{figure*}

\begin{figure*}
    \centering
    \includegraphics[trim=0 0 25mm 0, clip,width=0.8\linewidth]{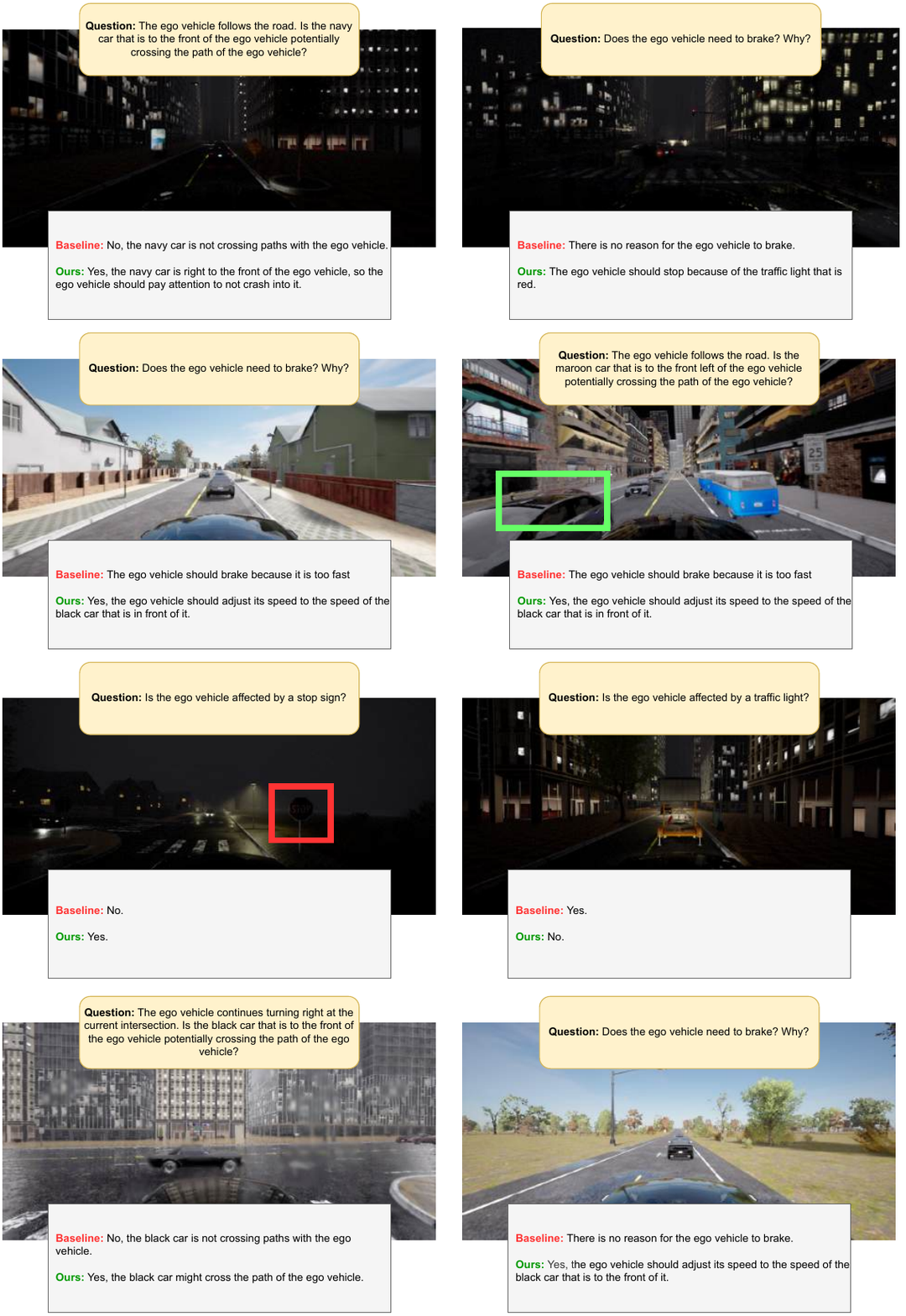}
    \caption{\textbf{Qualitative Results from DriveLM-CARLA test set.} These examples present a comparison between baseline and our results.}
    \label{fig:enter-label}
\end{figure*}

\end{document}